\documentclass{INTERSPEECH2023}
\usepackage{subfigure}
\usepackage{lipsum}
\interspeechcameraready

\title{AQUALLM: Audio Question Answering Data Generation Using Large Language Models}
\name{Swarup Ranjan Behera, Krishna Mohan Injeti, Jaya Sai Kiran Patibandla, \\Praveen Kumar Pokala, Balakrishna Reddy Pailla}
\address{Reliance Jio AICoE, Hyderabad, India}
\email{\{swarup.behera,injeti.mohan,jaya.patibandla,praveenkumar4.p,balakrishna.pailla\}@ril.com}

\begin{document}

\maketitle
 
\begin{abstract}
Audio Question Answering (AQA) constitutes a pivotal task in which machines analyze both audio signals and natural language questions to produce precise natural language answers. The significance of possessing high-quality, diverse, and extensive AQA datasets cannot be overstated when aiming for the precision of an AQA system. While there has been notable focus on developing accurate and efficient AQA models, the creation of high-quality, diverse, and extensive datasets for the specific task at hand has not garnered considerable attention. To address this challenge, this work makes several contributions. We introduce a scalable AQA data generation pipeline, denoted as the AQUALLM framework, which relies on Large Language Models (LLMs). This framework utilizes existing audio-caption annotations and incorporates state-of-the-art LLMs to generate expansive, high-quality AQA datasets. Additionally, we present three extensive and high-quality benchmark datasets for AQA, contributing significantly to the progression of AQA research. AQA models trained on the proposed datasets set superior benchmarks compared to the existing state-of-the-art. Moreover, models trained on our datasets demonstrate enhanced generalizability when compared to models trained using human-annotated AQA data. Code and datasets will be accessible on GitHub~\footnote{\url{https://github.com/swarupbehera/AQUALLM}}.
\end{abstract}

\section{Introduction}
\label{sec:intro}
Audio Question Answering (AQA) involves the task of generating natural language responses based on the content of audio signals. Similar to how humans seamlessly integrate auditory stimuli for a comprehensive understanding, an AQA system aims to provide accurate answers by deciphering relevant audio content in response to posed questions. In more formal terms, when presented with an audio signal, the AQA system endeavors to address natural language queries by discerning pertinent details within the auditory modality. For instance, consider an audio signal capturing a musical performance, with the question being, ``What instrument is playing?" To respond effectively, the AQA system must adeptly interpret the auditory cues within the signal, much like how humans naturally synthesize sensory information for a thorough understanding. This emphasizes the critical importance of the AQA system's ability to comprehend the connections between audio content and natural language queries to deliver precise responses.

The AQA system holds the potential to transform our interactions with audio content, enabling users to engage with audio-based systems in a seamless and productive manner~\cite{Clotho-AQA}. It serves various practical purposes, potentially transforming lives. It functions as a tool for interpreting natural language questions, providing accurate audio-based answers-particularly valuable for individuals with auditory impairments~\cite{DAQA}. Additionally, it assists professionals in diagnosing and analyzing medical audio data, thereby improving patient care and diagnostics. Beyond these applications, there are other potential use cases that underscore the system's versatility and broad impact.

For the development of the AQA system, it is essential that the system comprehends the contents of both audio signals and natural language questions. This necessitates the availability of numerous annotated audio-based datasets. To the best of our knowledge, there are currently five datasets for AQA: {\it ClothoAQA}~\cite{Clotho-AQA}, {\it DAQA}~\cite{DAQA}, {\it CLEAR}~\cite{CLEAR}, {\it mAQA}~\cite{behera23_interspeech}, and {\it Audio-MUSIC-AVQA}~\cite{Li2022LearningTA}. {\it ClothoAQA} is sourced from audio files in the Clotho dataset~\cite{CLOTHO}, originally designed for audio captioning, with a note on potential human biases and errors due to its crowd-sourced nature. Both {\it CLEAR} and {\it DAQA} datasets are algorithmically generated, featuring artificially created combinations of audio sequences, questions, and answers. However, these datasets often encounter challenges in representing diverse and natural scenes. {\it mAQA} is a multilingual dataset translated from {\it ClothoAQA} into eight languages, broadening its applicability and linguistic coverage. {\it Audio-MUSIC-AVQA} is a subset of the {\it MUSIC-AVQA}~\cite{Li2022LearningTA} dataset, specifically designed for scenarios involving music, providing insights into the AQA task limited to such contexts. For further details about these datasets, please refer to Table~\ref{tab:Newdata}.

The challenge in establishing robust AQA systems lies in the scarcity of large-scale, high-quality annotated AQA data, comprising the triplets: \textit{audio clip, question, and answer}. Manual annotation of these triplets is not only resource-intensive but also susceptible to human biases. Moreover, the current state-of-the-art AQA systems, trained on manually annotated data, do not attain human-level performance due to the lack of access to high-quality, diverse, large-scale AQA datasets. This motivates the exploration of a crucial question: {\it ``Can we develop a scalable pipeline to generate extensive, diverse, and high-quality annotated AQA datasets that can set new benchmarks surpassing existing state-of-the-art AQA baselines significantly?''}

\begin{table}[t!]
\caption{Using audio clips and their corresponding text captions, our AQUALLM framework, generates question-answer pairs, creating sizable AQA datasets with millions of examples, ideal for training AQA systems.}
\label{tab:Method}
\centering
\begin{tabular}{ll}
\toprule
\multicolumn{2}{c}{\textbf{Audio:} river\_mouth3.wav}\\
 \midrule
 \multicolumn{2}{c}{\includegraphics[width=0.905\linewidth]{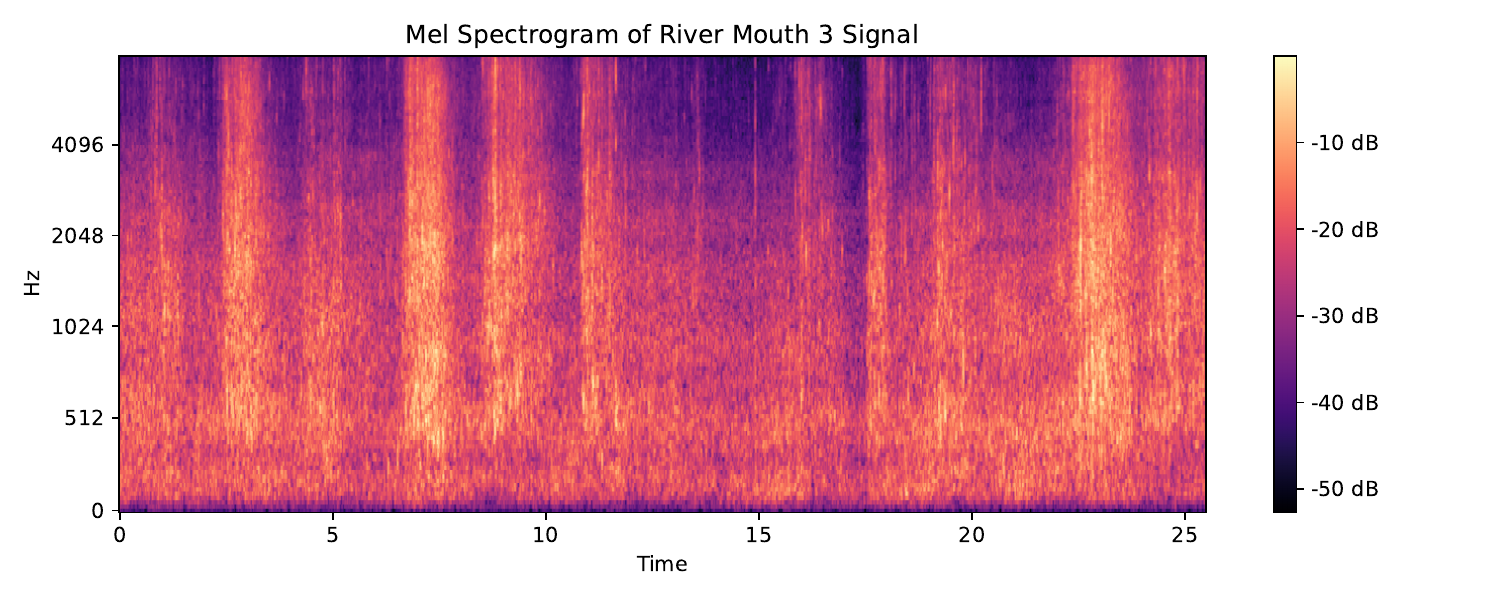}} \\
 \midrule
 \multicolumn{2}{c}{\textbf{Caption:} Waves of water are rolling against some rocks.}\\
 \midrule
 \multicolumn{2}{c}{$\downarrow$} \\
 \multicolumn{2}{c}{\framebox[1.1\width]{\textbf{AQUALLM}}}\\
 \multicolumn{2}{c}{$\downarrow$} \\
 \midrule
 \textbf{Question}   & \textbf{Answer}   \\
 \midrule
    % What is rolling against rocks?    &     Waves of water \\
    % What kind of water is rolling against rocks?    &     Waves \\
    What are waves of water rolling against? & Rocks \\
    Are waves rolling against rocks? & Yes \\
    % Is there a fancy alarm system that is constantly going off? & No \\
 \bottomrule
\end{tabular}
\end{table}

This work addresses the challenge of data scarcity in AQA by presenting several key contributions:

\begin{itemize}
    \item \textbf{AQUALLM Framework:} We introduce the AQUALLM framework, an automated AQA data generation pipeline. This scalable framework (refer to Table~\ref{tab:Method}) employs audio clips and associated text captions, incorporating state-of-the-art LLMs. It aims to streamline the creation of annotated AQA datasets, mitigating the manual effort and resource constraints typically associated with dataset annotation.
    \item \textbf{Annotated AQA Datasets:} Three large-scale AQA datasets - \textit{AQUALLM-AudioCaps}, \textit{AQUALLM-Clotho}, and \textit{AQUALLM-MACS} - are presented. Generated using the AQUALLM framework, these datasets offer expansiveness, high quality, and diverse content. They serve as valuable resources, addressing the shortage of annotated AQA data and contributing significantly to the advancement of AQA research.
    \item \textbf{New Benchmarks:} AQA models trained exclusively on the introduced datasets set new benchmarks. These benchmarks surpass existing state-of-the-art baselines, highlighting the effectiveness and superiority of models trained on our proposed datasets. This accomplishment represents a substantial progression in the field of AQA research.
\end{itemize}

In summary, our contributions encompass the development of an innovative automated AQUALLM framework, the creation of extensive and high-quality annotated AQA datasets, and the establishment of superior benchmarks. Together, these contributions enhance the landscape of AQA research.

\section{Related Work}
\label{sec:RelatedWork}
AQA is an emerging field and there is only five recorded instances of AQA research to the best of the authors' knowledge: {\it ClothoAQA}~\cite{Clotho-AQA}, {\it DAQA}~\cite{DAQA}, {\it CLEAR}~\cite{CLEAR}, {\it mAQA}~\cite{behera23_interspeech}, {\it Audio-MUSIC-AVQA}~\cite{Li2022LearningTA}, and {\it MWAFM}~\cite{li23v_interspeech}. There are only two available AQA models with source code in this field: {\it AquaNet}~\cite{Clotho-AQA} and {\it MWAFM}~\cite{li23v_interspeech}, with AquaNet being an LSTM-based model and MWAFM being an encoder-based model. Please note that the specifics of these models are outside the scope of this paper. Readers are encouraged to refer to the respective papers for a deeper understanding of these AQA models.

In the construction of any Question-Answer data generation pipeline, Question Generation (QG) plays a significant role. QG involves generating natural language questions based on various types of input. There are different categories of QG depending on the input source: Textual Question Generation (TQG)~\cite{TQG1,TQG2,TQG3} when the input is text, Visual Question Generation (VQG)~\cite{VQG1,VQG2} when the input is an image or video, and Audio Question Generation (AQG)~\cite{Clotho-AQA,CLEAR,DAQA} when the input is an audio signal. QG can be accomplished using three main approaches: crowd-sourced~\cite{Clotho-AQA}, template-based~\cite{CLEAR,DAQA,TQG2}, and neural network-based models~\cite{TQG3,VQG1,VQG2}. Neural network-driven QG is favored for its data-driven learning, contextual comprehension, adaptability, and capability to handle intricate structures, resulting in more natural and coherent questions.

\begin{figure*}[!t]
  \centering
  \includegraphics[width=\linewidth]{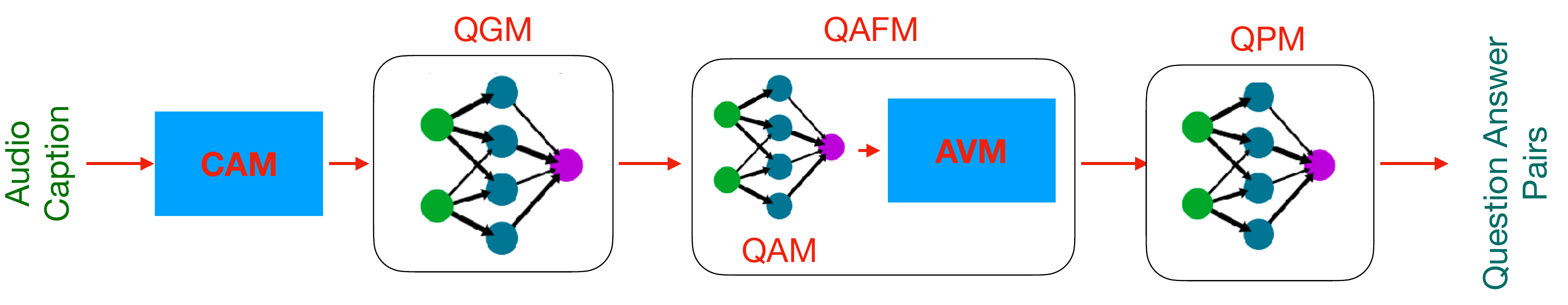}
  \caption{The key phases of AQUALLM Framework - CAM: Candidate Answer Extraction Module, QGM: Question Generation Module (LLM), QAFM: Question-Answer Filtering Module, which comprises QAM: Question-Answer Module (LLM) and AVM: Answer Verification Module, and QPM: Question Paraphrasing Module (LLM).}
  \label{fig:model}
\end{figure*}

\section{AQUALLM Framework}
\label{sec:SAAGF}
The AQUALLM framework begins with a set of audio-caption pairs (audio-captioning datasets), where each pair includes an audio clip and corresponding caption. The Candidate Answer Extraction Module (CAM) processes these pairs, automatically generating an initial set of potential answers from the captions following predefined rules. Subsequently, the Question Generation Module (QGM), powered by a Large Language Model (LLM), formulates a question for each potential answer, utilizing the caption as context. The ensuing Question-Answer Filtering Module (QAFM) validates each question-answer pair using an LLM. Successful pairs are then coupled with the corresponding audio clip. The final step involves the Question Paraphrasing Module (QPM), which paraphrases the generated questions to enhance variation. The outcome is the creation of a refined AQA example set: \textit{audio, question, answer}. This process transforms the initial audio-caption datasets into a comprehensive AQA dataset. Refer to Figure~\ref{fig:model} for a visual representation of this workflow, and detailed explanations of each module within the AQUALLM framework are provided below.

\subsection{Candidate Answer Extraction Module (CAM)}
\label{ssec:CandidateAnswerExtraction}

The process of candidate answer extraction covers a broad spectrum of answer types, including compound nouns, noun phrases, named entities, boolean answers, numbers, verbs, and their combinations, along with descriptive phrases. To extract potential answers from a given caption, we utilize spaCy\footnote{\url{https://spacy.io}} for parsing. Following this, we extract potential answer candidates based on Part-of-Speech (POS) and dependency parse tree annotations, as outlined in Table~\ref{tab:CAE}.

\begin{table}[h!]
      \caption{Categorization of answer candidates extraction mechanism into ICAC (In Caption Answer Candidates) and OCAC (Out of Caption Answer Candidates).}
      \label{tab:CAE}
      \centering
      % \resizebox{0.5\textwidth}{!}{
        \begin{tabular}{l p{0.735\linewidth}}
        \toprule
        \textbf{ICAC} & \textbf{Extract} \\
        \midrule 
        Nouns & Noun phrases, including named entities. \\
        Verbal & Verbal phrases that start with a verb and end with a verb/noun with open class tags in between. \\
        Adjective & Adjective phrases and sequences that start with an adjective and end with an adjective/noun with open class tags in between. \\
        Adverbial & Adverbial phrases and sequences that start with an adverb and end with an adverb/noun with open class tags in between. \\
        Cardinal & Numbers and cardinal phrases. \\
        \midrule
        \textbf{OCAC} & \textbf{Rules} \\
        \midrule
        Boolean & `Yes' and `no' are typically absent in captions, so we incorporate them as candidates and produce a question for each.\\
        Zero (0) & Due to the absence of `zero' object counts in captions, we insert a randomly sampled `How many?' question from another caption and modify the answer to  `zero'.\\
        \bottomrule
        \end{tabular}
        % }
\end{table}

\subsection{Question Generation Module (QGM)}
\label{ssec:QuestionGeneration}
The process of question generation involves taking a caption and an answer candidate as input, ultimately resulting in the generation of a question. The system is crafted to generate questions even when the answer doesn't precisely match the content of the caption. This flexibility enables the creation of questions for diverse types of answers, encompassing yes/no responses or situations where the answer is zero. Depending on the type of answer (Cf.~Table~\ref{tab:CAE}), we employ two distinct approaches for question generation. 

\begin{itemize}
    \item We employ a Large Language Model (LLM), specifically the T5 model~\cite{t5}, fine-tuned on the SQuAD1.1 dataset~\cite{squad1}, to generate questions for answer phrases categorized as nouns, verbs, adjectives, adverbials, cardinals, and `yes.' 
    \item For answer phrases `no' and `zero', we incorporate a randomly selected `How many?' question from another caption.
\end{itemize}

Although not fine-tuned for captions, this model, trained on question-answering data, handles captions well, as seen in our manual review of hundreds of questions.

\subsection{Question-Answer Filtering Module (QAFM)}
\label{ssec:QuestionAnswerFiltering}
Question generation models can sometimes generate content that deviates from the input source, known as `hallucination.' To address this issue, we ensure round-trip consistency by using a Question-Answer Module (QAM) to respond to the generated question based on the caption text and an Answer Verification Module (AVM) to validate the question-answer pairs.

\begin{itemize}
    \item \textbf{Question-Answer Module (QAM):} For answering questions, we employ a T5 model~\cite{t5}, fine-tuned on the SQuAD2.0 dataset~\cite{squad2}.

    \item \textbf{Answer Verification Module (AVM):} 
    If the answer provided by the QAM doesn't align with the answer candidate used during question generation (as described in Section~\ref{ssec:QuestionGeneration}), we reject the generated question. To determine the match between the candidate answer and the QA model's answer, AVM employs a token-level $F1$ score. If the score exceeds a predefined threshold ($0.55$ in this case), we consider them a match. A higher threshold indicates higher question quality. This answer verification contributes to maintaining the quality and consistency of generated questions and answers.
\end{itemize}

\subsection{Question Paraphrasing Module (QPM)}
\label{ssec:QuestionParaphrasing}
The QPM functions by taking a question as input and generating similar questions through rephrasing the original question. This approach enriches the adaptability of the AQA model and expands its comprehension by exposing it to a diverse set of examples. For question paraphrasing, we employ a T5 model~\cite{t5}, fine-tuned on the Quora Question Pair dataset~\cite{QQP} to generate similar questions. Specifically, for each question, we select the top 5 similar questions. However, we acknowledge the need for caution to prevent paraphrasing from introducing ambiguity or complexity into questions.

% \begin{table}[t!]
%       \caption{Comparative overview of AQA datasets. Number of audios (\#Audio) samples. Question-answer source (S) indicates whether questions and answers are crowd-sourced (C) or generated programmatically (P). Language (L) of the question and answer: English (E). Number of unique questions (\#Q). Number of unique answers (\#A).}
%       \label{tab:AQAdata}
%       \centering
%         \begin{tabular}{lccccc}
%         \toprule
%         \textbf{Dataset} & \textbf{\#Audios} &  \textbf{S} & \textbf{L} & \textbf{\#Q} & \textbf{\#A} \\
%         \midrule
%         {\it ClothoAQA}~\cite{Clotho-AQA}	& $1991$ & C      & E         & $9153$ & $830$ \\
%         {\it CLEAR}~\cite{CLEAR}	& $50000$ & P        & E       & $130957$  & $47$ \\
%         {\it DAQA}~\cite{DAQA} & $100000$  & P & E             & $599294$  & $36$ \\
%         % MUSIC-AVQA~\cite{Li2022LearningTA} & $93000$  & P & E             & 599294  & 36 \\
%         \bottomrule
%         \end{tabular}
% \end{table}

\section{Experimental Results}
\label{sec:Experiments}
We assess the performance of the proposed AQUALLM framework through the following steps: (i) Creation of AQA datasets. (ii) Training a state-of-the-art AQA model using the proposed datasets and comparing its performance with the same model trained on current AQA datasets.

\subsection{AQA Dataset Creation}
\label{sec:SAAGFDataset}
To evaluate the effectiveness of AQUALLM framework, we utilized it to generate AQA triplets from the audio captions sourced from three audio captioning datasets such as {\it AudioCaps}, {\it Clotho}, and {\it MACS}. Additional details about these datasets are available in Table~\ref{tab:ACdata}. Details for the datasets generated through the AQUALLM framework-namely, \textit{AQUALLM-AudioCaps, AQUALLM-Clotho, and AQUALLM-MACS}, are outlined in Table~\ref{tab:Newdata}. For a comprehensive comparison, we have also incorporated details of the current AQA datasets.

\begin{table}[t!]
      \caption{Comparative overview of Audio Captioning datasets. Number of audios (\#A). Audio duration (AD) in seconds. Number of captions per audio (C/A). Number of captions (\#C).  Vocab size (V).}
      \label{tab:ACdata}
      \centering
      % \resizebox{0.5\textwidth}{!}{
        \begin{tabular}{lccccc}
        \toprule
        \textbf{Dataset} & \textbf{\#A} &  \textbf{AD} & \textbf{C/A} & \textbf{\#C} & \textbf{V} \\
        \midrule
        {\it Clotho}	& $5929$ & $15-30$ & $5$ & $19195$ & $4365$ \\
        {\it AudioCaps}	& $51308$ & $10$ & $1,5$ & $49838$ & $5066$ \\
        {\it MACS} & $3930$  & $10$ & $2-5$ & $17275$ & $2776$ \\
        % AudioCaption~\cite{} & 3930  & TAU UAC & 17275             & E  & 2-5 \\
        \bottomrule
        \end{tabular}
        % }
\end{table}

\begin{table}[t!]
          \caption{Comparative overview of current AQA datasets and \textbf{proposed AQA datasets}. Number of audios (\#A). Language (L) of the question and answer: English (E). Number of unique questions (\#Q). Number of unique answers (\#A).}
      \label{tab:Newdata}
      \centering
      % \resizebox{0.5\textwidth}{!}{
        \begin{tabular}{lcccc}
        \toprule
        \textbf{Dataset} & \textbf{\#A} &  \textbf{L} & \textbf{\#Q} & \textbf{\#Ans} \\
        \midrule
        {ClothoAQA}	& $1991$ & E         & $9153$ & $830$ \\
        {CLEAR}	& $50000$ & E       & $130957$  & $47$ \\
        {DAQA} & $100000$  & E             & $599294$  & $36$ \\
        {\bf AQUALLM-Clotho}	& $5929$ & E       & $438600$  & $25235$ \\
        {\bf AQUALLM-AudioCaps}	& $51308$ & E         & $728310$ & $35008$ \\
        {\bf AQUALLM-MACS} & $3930$  & E             & $268875$ & $11874$ \\
        % MUSIC-AVQA~\cite{Li2022LearningTA} & 93000  & P & E             & 599294  & 36 \\
        \bottomrule
        \end{tabular}
        % }
\end{table}

\subsection{AQA Model Training and Comparison}
In line with existing literature~\cite{Clotho-AQA}, we consider AQA as a classification task where answers are treated as labels, and they can consist of `yes' and `no' (binary classifications) or single or multiple words, such as `eleven' and `flowing water' (multi-class classifications). There are only two AQA models with available source code: AquaNet~\cite{Clotho-AQA} and MWAFM~\cite{li23v_interspeech}. Additionally, MWAFM has demonstrated superiority over AquaNet. The experimental setup for the AQA task remains consistent with the one utilized in MWAFM.

The performance of the MWAFM model is displayed in Table~\ref{tab:result1}, illustrating its outcomes on both the current dataset (ClothoAQA) and the proposed datasets. Notably, the accuracy, the chosen evaluation metric, exhibits a significant improvement when training the MWAFM model on the proposed dataset (AQUALLM-Clotho) compared to the existing dataset (ClothoAQA). Both datasets are derived from the Clotho audio-captioning dataset. One can observe that the accuracies on the proposed datasets exceed $95\%$, substantiating the fact that the proposed datasets are of very high quality.

\begin{table}[!t]
  \caption{Comparision of accuracies (\%) of MWAFM binary  classifier on existing dataset and \textbf{proposed datasets}.}
  \label{tab:result1}
  \centering
  % \resizebox{0.35\textwidth}{!}{
    \begin{tabular}{lc}
    \toprule
     \textbf{Dataset} & \textbf{MWAFM Accuracy} \\
     \midrule
        {\it ClothoAQA}   &  $68.75$ \%\\
        {\bf AQUALLM-Clotho}     &  $95.58$ \% \\
        {\bf AQUALLM-AudioCaps}   &  $95.86$ \% \\
        {\bf AQUALLM-MACS}     &  $96.84$ \%\\
     \bottomrule
    \end{tabular}
    % }
\end{table}

\section{Conclusion and Future Work}
\label{sec:Conclusion}
In conclusion, the AQUALLM framework marks a significant advance in AQA research. Built on LLMs, it efficiently generates extensive, high-quality AQA datasets by leveraging existing audio-caption annotations. The framework introduces three benchmark datasets that outperform existing benchmarks, addressing the scarcity of human-annotated AQA data.

Looking forward, there are promising paths for exploration. Our decision to defer multi-class classification ensures focus on clarity and precision in addressing specific challenges. Future studies may delve into multi-class classification for a deeper understanding of AQA systems. The intentional choice of the T5 model prompts ongoing efforts to develop a Python package, fostering adaptability with different LLMs. A key future focus involves a thorough examination of the quality of AQA data generated by AQUALLM. This includes assessing question-context alignment, ensuring answer accuracy, and avoiding unintentional biases. Developing reliable methods to measure AQA dataset quality is essential for refining the AQUALLM framework, furthering its role in advancing AQA research.

\bibliographystyle{IEEEtran}
\bibliography{mybib}

\begin{thebibliography}{10}
\providecommand{\url}[1]{#1}
\csname url@samestyle\endcsname
\providecommand{\newblock}{\relax}
\providecommand{\bibinfo}[2]{#2}
\providecommand{\BIBentrySTDinterwordspacing}{\spaceskip=0pt\relax}
\providecommand{\BIBentryALTinterwordstretchfactor}{4}
\providecommand{\BIBentryALTinterwordspacing}{\spaceskip=\fontdimen2\font plus
\BIBentryALTinterwordstretchfactor\fontdimen3\font minus \fontdimen4\font\relax}
\providecommand{\BIBforeignlanguage}[2]{{%
\expandafter\ifx\csname l@#1\endcsname\relax
\typeout{** WARNING: IEEEtran.bst: No hyphenation pattern has been}%
\typeout{** loaded for the language `#1'. Using the pattern for}%
\typeout{** the default language instead.}%
\else
\language=\csname l@#1\endcsname
\fi
#2}}
\providecommand{\BIBdecl}{\relax}
\BIBdecl

\bibitem{Clotho-AQA}
S.~Lipping, P.~Sudarsanam, K.~Drossos, and T.~Virtanen, ``Clotho-{AQA}: {A} crowdsourced dataset for audio question answering,'' \emph{in proceedings of the IEEE International European Signal Processing Conference}, pp. 1140--1144, 2022.

\bibitem{DAQA}
H.~M. Fayek and J.~Johnson, ``Temporal reasoning via audio question answering,'' \emph{IEEE Transactions on Audio, Speech, and Language Processing}, vol.~28, pp. 2283--2294, 2020.

\bibitem{CLEAR}
J.~Abdelnour, G.~Salvi, and J.~Rouat, ``Clear: A dataset for compositional language and elementary acoustic reasoning,'' 2019.

\bibitem{behera23_interspeech}
S.~R. Behera, B.~R. Pailla, A.~M. Tripathi, M.~B. Rathod, and T.~Karavadi, ``{Towards Multi-Lingual Audio Question Answering},'' in \emph{proceedings of the IEEE International Speech Communication Association}, pp. 356--360, 2023.

\bibitem{Li2022LearningTA}
G.~Li, Y.~Wei, Y.~Tian, C.~Xu, J.~Wen, and D.~Hu, ``Learning to answer questions in dynamic audio-visual scenarios,'' \emph{in the proceedings of the International Conference on Computer Vision and Pattern Recognition}, pp. 19\,086--19\,096, 2022.

\bibitem{CLOTHO}
K.~Drossos, S.~Lipping, and T.~Virtanen, ``Clotho: an audio captioning dataset,'' in \emph{proceeding of the IEEE International Conference on Acoustics, Speech and Signal Processing}, pp. 736--740, 2020.

\bibitem{li23v_interspeech}
G.~Li, Y.~Xu, and D.~Hu, ``{Multi-Scale Attention for Audio Question Answering},'' in \emph{proceedings of the IEEE International Speech Communication Association}, 2023, pp. 3442--3446, 2023.

\bibitem{TQG1}
R.~Puri, R.~Spring, M.~Shoeybi, M.~Patwary, and B.~Catanzaro, ``Training question answering models from synthetic data,'' in \emph{proceedings of the International Conference on Empirical Methods in Natural Language Processing}, pp. 5811--5826, 2020.

\bibitem{TQG2}
C.~Lyu, L.~Shang, Y.~Graham, J.~Foster, X.~Jiang, and Q.~Liu, ``Improving unsupervised question answering via summarization-informed question generation,'' in \emph{proceedings of the International Conference on Empirical Methods in Natural Language Processing}, pp. 4134--4148, 2021.

\bibitem{TQG3}
S.~Narayan, G.~Sim{\~a}es, J.~Ma, H.~Craighead, and R.~T. McDonald, ``Qurious: Question generation pretraining for text generation,'' \emph{ArXiv}, vol. abs/2004.11026, 2020.

\bibitem{VQG1}
A.~Akula, S.~Changpinyo, B.~Gong, P.~Sharma, S.-C. Zhu, and R.~Soricut, ``{C}ross{VQA}: Scalably generating benchmarks for systematically testing {VQA} generalization,'' in \emph{proceedings of the Intrenational Conference on Empirical Methods in Natural Language Processing}, pp. 2148--2166, 2021.

\bibitem{VQG2}
S.~Changpinyo, D.~Kukliansy, I.~Szpektor, X.~Chen, N.~Ding, and R.~Soricut, ``All you may need for {VQA} are image captions,'' in \emph{proceedings of the International Conference of the North American Chapter of the Association for Computational Linguistics: Human Language Technologies}, pp. 1947--1963, 2022.

\bibitem{t5}
A.~Roberts, C.~Raffel, K.~Lee, M.~Matena, N.~Shazeer, P.~J. Liu, S.~Narang, W.~Li, and Y.~Zhou, ``Exploring the limits of transfer learning with a unified text-to-text transformer,'' Google, Tech. Rep., 2019.

\bibitem{squad1}
P.~Rajpurkar, J.~Zhang, K.~Lopyrev, and P.~Liang, ``Squad: 100,000+ questions for machine comprehension of text,'' in \emph{proceedings of the International Conference on Empirical Methods in Natural Language Processing}, 2016.

\bibitem{squad2}
P.~Rajpurkar, R.~Jia, and P.~Liang, ``Know what you don’t know: Unanswerable questions for squad,'' in \emph{Annual Meeting of the Association for Computational Linguistics}, 2018.

\bibitem{QQP}
Z.~Chen, H.~Zhang, X.~Zhang, and L.~Zhao, ``Quora question pairs,'' 2017.

\end{thebibliography}

\end{document}